%% file: main.tex
\documentclass[10pt,twocolumn,letterpaper]{article}

\usepackage{iccv}
\usepackage{times}
\usepackage{epsfig}
\usepackage{graphicx}
\usepackage{amsmath}
\usepackage{amssymb}
\usepackage{booktabs}
\usepackage{multirow}
\usepackage{subcaption}
\usepackage[inline]{enumitem}

\usepackage{booktabs}
\usepackage{multicol}
\usepackage{color}
\usepackage{caption}
\usepackage{hhline}
\usepackage{pifont}
\usepackage{threeparttable}
\usepackage{makecell}
\usepackage{algorithm} 
\usepackage{listings}
\usepackage{amsfonts,amssymb}
\usepackage{graphicx}
\usepackage{tabularx}
\usepackage{lipsum}
\newcommand\blfootnote[1]{%
  \begingroup
  \renewcommand\thefootnote{}\footnote{#1}%
  \addtocounter{footnote}{-1}%
  \endgroup
}

\usepackage[pagebackref=true,breaklinks=true,letterpaper=true,colorlinks,bookmarks=false]{hyperref}

\iccvfinalcopy % *** Uncomment this line for the final submission

\def\iccvPaperID{9078} % *** Enter the ICCV Paper ID here

\ificcvfinal\fi
\usepackage[capitalize]{cleveref}
\crefname{section}{Sec.}{Secs.}
\Crefname{section}{Section}{Sections}
\Crefname{table}{Table}{Tables}
\crefname{table}{Tab.}{Tabs.}

\usepackage[accsupp]{axessibility}  % Improves PDF readability for those with disabilities.

\begin{document}

%%%%%%%%% TITLE
\title{FastViT: A Fast Hybrid Vision Transformer \\using Structural Reparameterization}

\author{
Pavan Kumar Anasosalu Vasu$^\dagger$ \quad James Gabriel \quad Jeff Zhu \quad Oncel Tuzel \quad Anurag Ranjan$^\dagger$ \\ \\ Apple
}

\maketitle

\setlength{\abovedisplayskip}{3pt}
\setlength{\belowdisplayskip}{3pt}

%%%%%%%%% ABSTRACT
\begin{abstract}

The recent amalgamation of transformer and convolutional designs has led to steady improvements in accuracy and efficiency of the models. 
In this work, we introduce {FastViT}, a hybrid vision transformer architecture that obtains the state-of-the-art latency-accuracy trade-off. To this end, we introduce a novel token mixing operator, RepMixer, a building block of FastViT, that uses structural reparameterization to lower the memory access cost by removing  skip-connections in the network. We further apply train-time overparametrization and large kernel convolutions to boost accuracy and empirically show that these choices have minimal effect on latency. We show that -- our model is 3.5$\times$ faster than CMT, a recent state-of-the-art hybrid transformer architecture, 4.9$\times$ faster than EfficientNet, and 1.9$\times$ faster than ConvNeXt on a mobile device for the same accuracy on the ImageNet dataset. At similar latency, our model obtains 4.2\% better Top-1 accuracy on ImageNet than MobileOne. Our model consistently outperforms competing architectures across several tasks -- image classification, detection, segmentation and 3D mesh regression with significant improvement in latency on both a mobile device and a desktop GPU. Furthermore, our model is highly robust to out-of-distribution samples and corruptions, improving over competing robust models. Code and models are available at \url{https://github.com/apple/ml-fastvit} \blfootnote{corresponding authors: {\{panasosaluvasu, anuragr\}@apple.com}}

\end{abstract}

%%%%%%%%% BODY TEXT
\input{introduction}
\input{related_work}

\input{method}
\input{experiments}

\input{conclusion}

\paragraph{Acknowledgements}
We thank Jen-Hao Rick Chang, Skyker Seto and Hadi Pouransari for helpful discussions and reviewing the manuscript.

{\small
\bibliographystyle{ieee_fullname}
\bibliography{egbib}
}

\input{appendix}

\end{document}

%% file: introduction.tex
\section{Introduction}
\label{sec:introduction}

\begin{figure*}[t]
  \centering
  \begin{subfigure}{.48\linewidth}
    \centering
    \includegraphics[width=\linewidth]{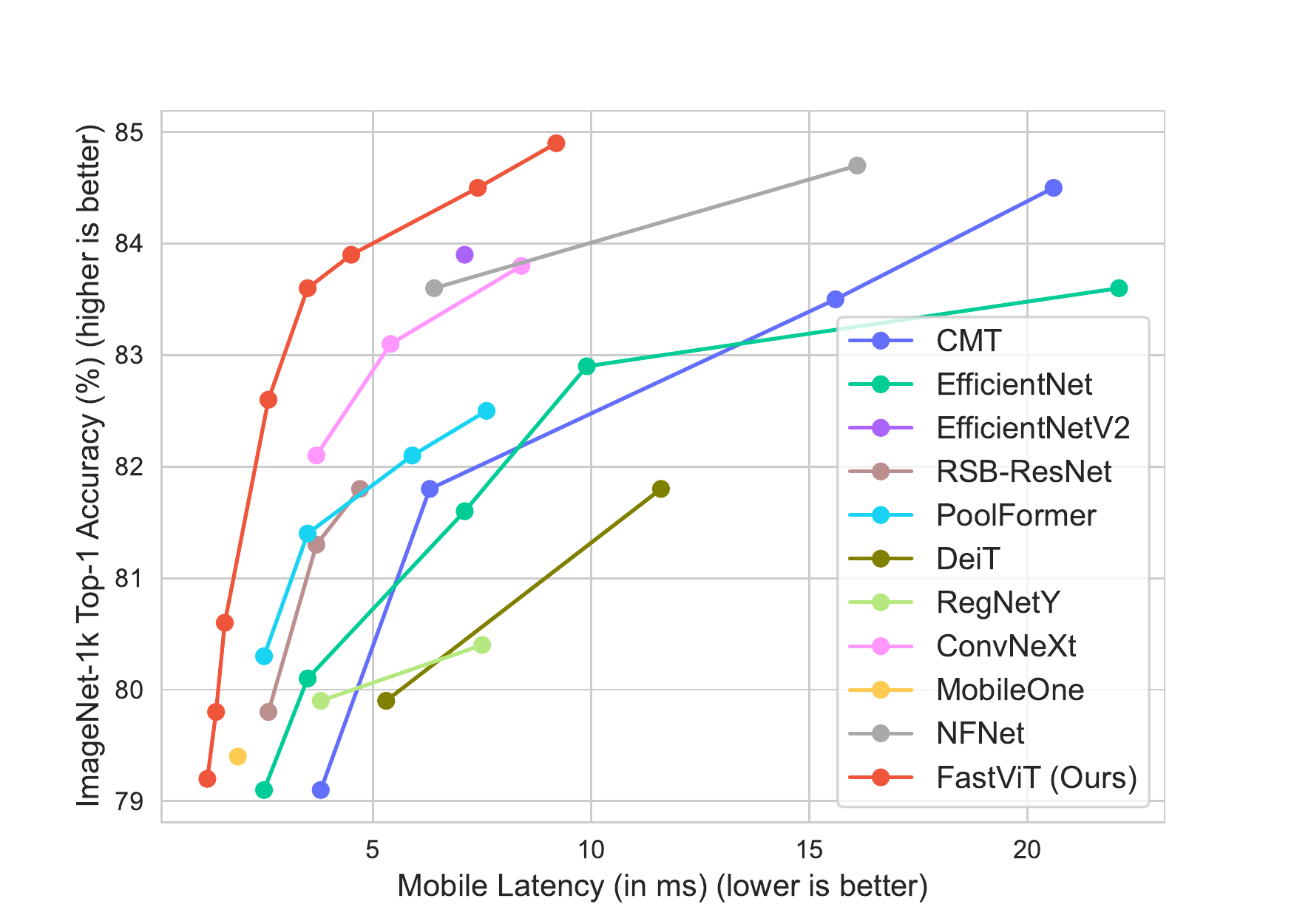}
    \caption{}
  \end{subfigure}%
  \hfill
  \begin{subfigure}{.48\linewidth}
    \centering
    \includegraphics[width=\linewidth]{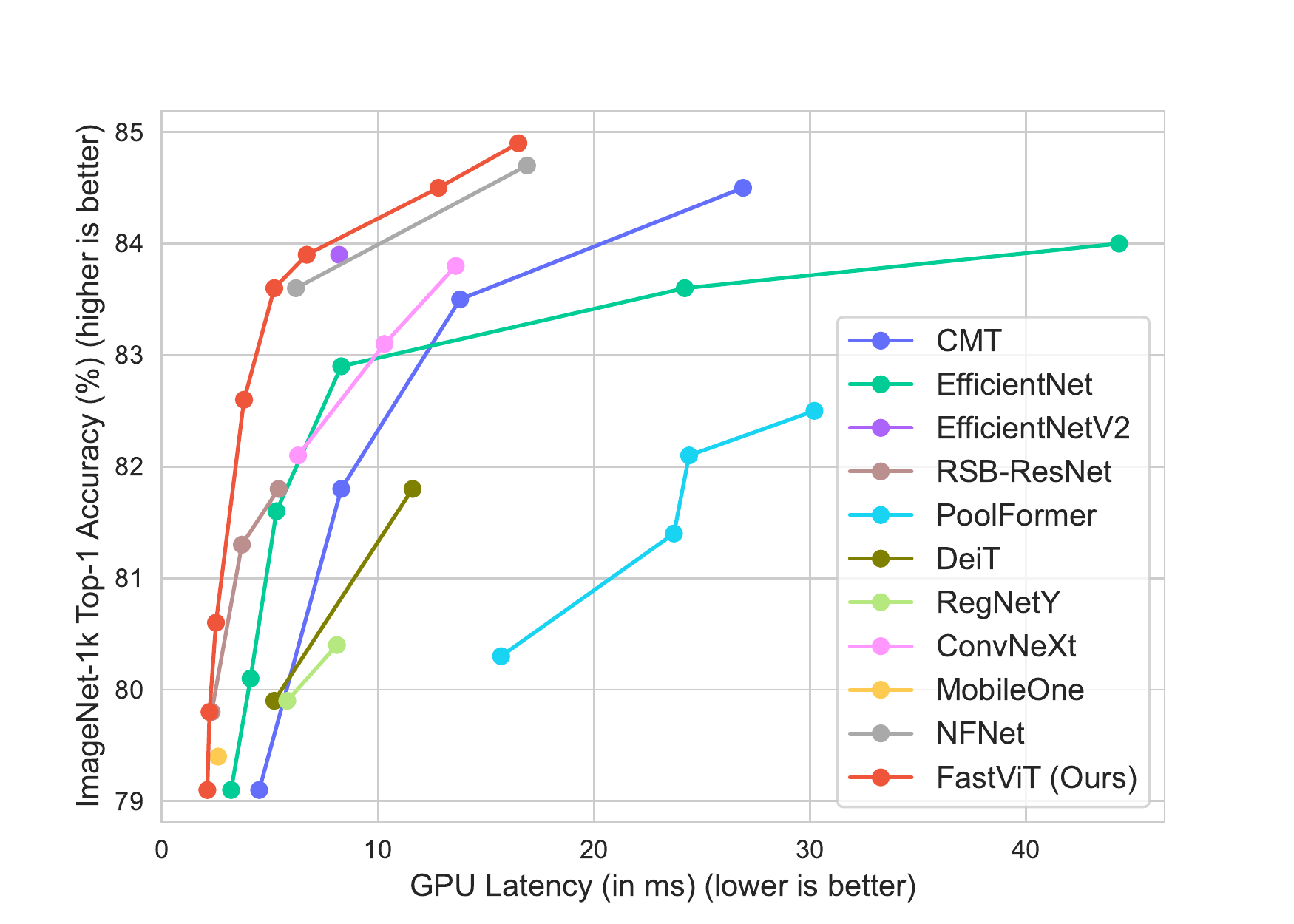}
    \caption{}
  \end{subfigure}%
    \vspace{-2mm}
  \caption{(a) Accuracy vs. Mobile latency scaling curves of recent methods. The models are benchmarked on an iPhone 12 Pro following ~\cite{mobileone}. (b) Accuracy vs. GPU latency scaling curves of recent methods. For better readability only models with Top-1 accuracy better than 79\% are plotted. See supplementary materials for more plots.
  Across both compute fabrics, our model has the best accuracy-latency tradeoff.}
  \label{fig:teaser}
  \vspace{-4mm}
\end{figure*}

Vision Transformers~\cite{ViT} have achieved state-of-the-art performance on several tasks such as image classification, detection and segmentation~\cite{Swin}. However, these models have traditionally been computationally expensive. Recent works ~\cite{zhai2021aft, marin2021token, litv2, linformer, reformer} have proposed methods to lower the compute and memory requirements of vision transformers. Recent hybrid architectures~\cite{litv2, Cmt, d2021convit, cvt} effectively combine the strengths of convolutional architectures and transformers to build  architectures that are highly competitive on a wide range of computer vision tasks. Our goal is to build a model that achieves state-of-the-art latency-accuracy trade-off.

Recent vision and hybrid transformer models~\cite{deit, Cmt, litv1, litv2} follow the Metaformer~\cite{metaformer} architecture, which consists of a token mixer with a skip connection followed by Feed Forward Network (FFN) with another skip connection. These skip connections account for a significant overhead in latency due to increased memory access cost~\cite{Ding_2021_repvgg, mobileone}.
To address this latency overhead, we introduce ~\textit{RepMixer}, a fully reparameterizable token mixer that uses structural reparameterization to remove the skip-connections.
The RepMixer block also uses depthwise convolutions for spatial mixing of information similar to ConvMixer~\cite{trockman2022convmixer}. However, the key difference is that our module can be reparameterized at inference to remove any branches. 

To further improve on latency, FLOPs and parameter count, we replace all dense k$\times$k convolutions with their factorized version, i.e. depthwise followed by pointwise convolutions. This is a common approach used by efficient architectures~\cite{MobileNet_v1, MobileNet_v2, Mobilenet_v3} to improve on efficiency metrics, but, naively using this approach hurts performance as seen in Table~\ref{tab:ablation_arch_choices}. In order to increase capacity of the these layers, we use linear train-time overparameterization as introduced in~\cite{Ding_2021_repvgg, Ding_2019_ICCV, ding2021diverse, mobileone, NEURIPS2020_expandnets}. These additional branches are only introduced during training and are reparameterized at inference.  

In addition, we use large kernel convolutions 
in our network. This is because, although self-attention based token mixing is highly effective to attain competitive accuracy, they are inefficient in terms of latency~\cite{marin2021token}.
Therefore, we incorporate large kernel convolutions in Feed Forward Network (FFN)~\cite{ViT} layer and patch embedding layers. These changes have minimal impact on overall latency of the model while improving performance.  

Thus, we introduce \emph{FastViT} that is based on three key design principles-- i) use of RepMixer block to remove skip connections, ii) use of linear train-time overparameterization to improve accuracy, iii) use of large convolutional kernels to substitute self-attention layers in early stages.

{FastViT} achieves significant improvements in latency compared to other hybrid vision transformer architectures while maintaining accuracy on several tasks like -- image classification, object detection, semantic segmentation and 3d hand mesh estimation. We perform a comprehensive analysis by deploying recent state-of-the-art architectures on an iPhone 12 Pro device and an NVIDIA RTX-2080Ti desktop GPU.

In Figure~\ref{fig:teaser}, we show that, at ImageNet Top-1 accuracy of 83.9\%, our model is 4.9$\times$ faster than EfficientNet-B5~\cite{EfficientNet}, 1.6$\times$ faster than EfficientNetV2-S~\cite{efficientnet_v2_quoc}, 3.5$\times$ faster than CMT-S~\cite{Cmt} and 1.9$\times$ faster than ConvNeXt-B~\cite{ConvNext} on an iPhone 12 Pro mobile device. At ImageNet Top-1 accuracy of 84.9\% our model is just as fast as NFNet-F1~\cite{nfnets} on a desktop GPU while being 66.7\% smaller, using 50.1\% less FLOPs and 42.8\% faster on mobile device.
At latency of 0.8ms on an iPhone 12 Pro mobile device, our model obtains 4.2\% better Top-1 accuracy on ImageNet than MobileOne-S0. 
For object detection and instance segmentation on MS COCO using Mask-RCNN~\cite{Mask_RCNN} head, our model attains comparable performance to CMT-S~\cite{Cmt} while incurring 4.3$\times$ lower backbone latency. For semantic segmentation on ADE20K, our model improves over PoolFormer-M36~\cite{metaformer} by 5.2\%, while incurring a 1.5$\times$ lower backbone latency on an iPhone 12 Pro mobile device. On 3D hand mesh estimation task, our model is 1.9$\times$ faster than MobileHand~\cite{mobilehand} and 2.8$\times$ faster than recent state-of-the-art MobRecon~\cite{mobrecon} when benchmarked on GPU. 

% \\

In addition to accuracy metrics, we also study the robustness of our models to corruption and out-of-distribution samples which does not always correlate well with accuracy.
For example, PVT~\cite{PVT_v1} achieves highly competitive performance on ImageNet dataset, but has very poor robustness to corruption and out-of-distribution samples as reported in Mao et al.~\cite{mao2022robust}.
In real world applications, 
using a robust model in such a scenario can significantly improve user experience.   
We demonstrate the robustness of our architecture on popular benchmarks and show that our models are highly robust to corruption and out-of-distribution samples while being significantly faster than competing robust models. 
% \vspace{-0.2cm}
In summary, our contributions are as follows:
\begin{itemize}[leftmargin=*]
\setlength\itemsep{-.1em}
    \item We introduce \textit{FastViT}, a hybrid vision transformer that uses structural reparameterization to obtain lower memory access cost and increased capacity, achieving state-of-the-art accuracy-latency trade-off.

    \item We show that our models are the fastest in terms of latency on two widely used platforms -- mobile device and desktop GPU.
    \item We show that our models generalize to many tasks -- image classification, object detection, semantics segmentation, and 3D hand mesh regression.
    \item We show that our models are robust to corruption and out-of-distribution samples and significantly faster than competing robust models.
\end{itemize}

%% file: related_work.tex
\section{Related Work}
\label{sec:related_work}

For the past decade, convolutional neural networks have been the standard architecture for vision models~\cite{resnet, RegNet, ResNeXt, EfficientNet, efficientnet_v2_quoc, MobileNet_v1, MobileNet_v2, Mobilenet_v3, ConvNext, Ding_2021_repvgg, mobileone}. More recently, transformers have shown great success on computer vision tasks~\cite{ViT, vitc_arxiv, deit, Swin, PVT_v1, cait, litv1, litv2}. Unlike convolutional layers, the self-attention layers in vision transformers provide a global context by modeling long-range dependencies. Unfortunately, this global scope often comes at a high computational price~\cite{marin2021token}. Works like~\cite{litv2, linformer, reformer, marin2021token} address ways to alleviate computational costs associated with self-attention layers. In our work, we explore an efficient alternative to self-attention layers for lower latency.

\vspace{-3.5mm}
\paragraph{Hybrid Vision Transformers} In order to design efficient networks while maintaining accuracy,
recent works introduce hybrid architectures that combine convolutional and transformer design to effectively capture local and global information.
Some designs replace the patchify stem~\cite{ViT} with convolutional layers~\cite{vitc_arxiv}, introduce early convolutional stages~\cite{CoAtNet,litv2} or implicitly hybridize through windowed attention~\cite{Swin, Twins}. More recent works build explicit hybrid structures for better exchange of information between tokens (or patches)~\cite{Cmt, d2021convit, cvt}. In majority of the hybrid architectures, token mixers are predominantly self-attention based. Recently, MetaFormer~\cite{metaformer} introduced Pooling, a simple and efficient candidate for token mixing.

\vspace{-3.5mm}
\paragraph{Structural Reparameterization} Recent work~\cite{Ding_2021_repvgg, mobileone} shows the benefits of reparameterizing skip connections to lower memory access cost. In our model, we introduce a novel architectural component \textit{RepMixer}, that is reparameterizable at inference. For better efficiency, works like~\cite{MobileNet_v1, MobileNet_v2, Mobilenet_v3, ShuffleNet_v2, ShuffleNet} introduce factorized k$\times$k convolutions using depthwise or grouped convolutions followed by 1$\times$1 pointwise convolutions. While this approach is highly effective in improving the overall efficiency of the model, the lower parameter count can lead to reduced capacity. Recently, linear train-time over overparameterization was introduced in~\cite{mobileone, Ding_2019_ICCV, ding2021diverse} to improve capacity of such models. We use factorized k$\times$k convolutions in our model and boost the capacity of these layers using linear train-time overparameterization. To the best of our knowledge, structural reparameterization to remove skip connections and linear overparameterization has not been attempted in any prior hybrid transformer architecture.

%% file: method.tex
\section{Architecture} \label{sec:arch}

\begin{figure*}[t]
    \centering
    \includegraphics[width=0.97\linewidth]{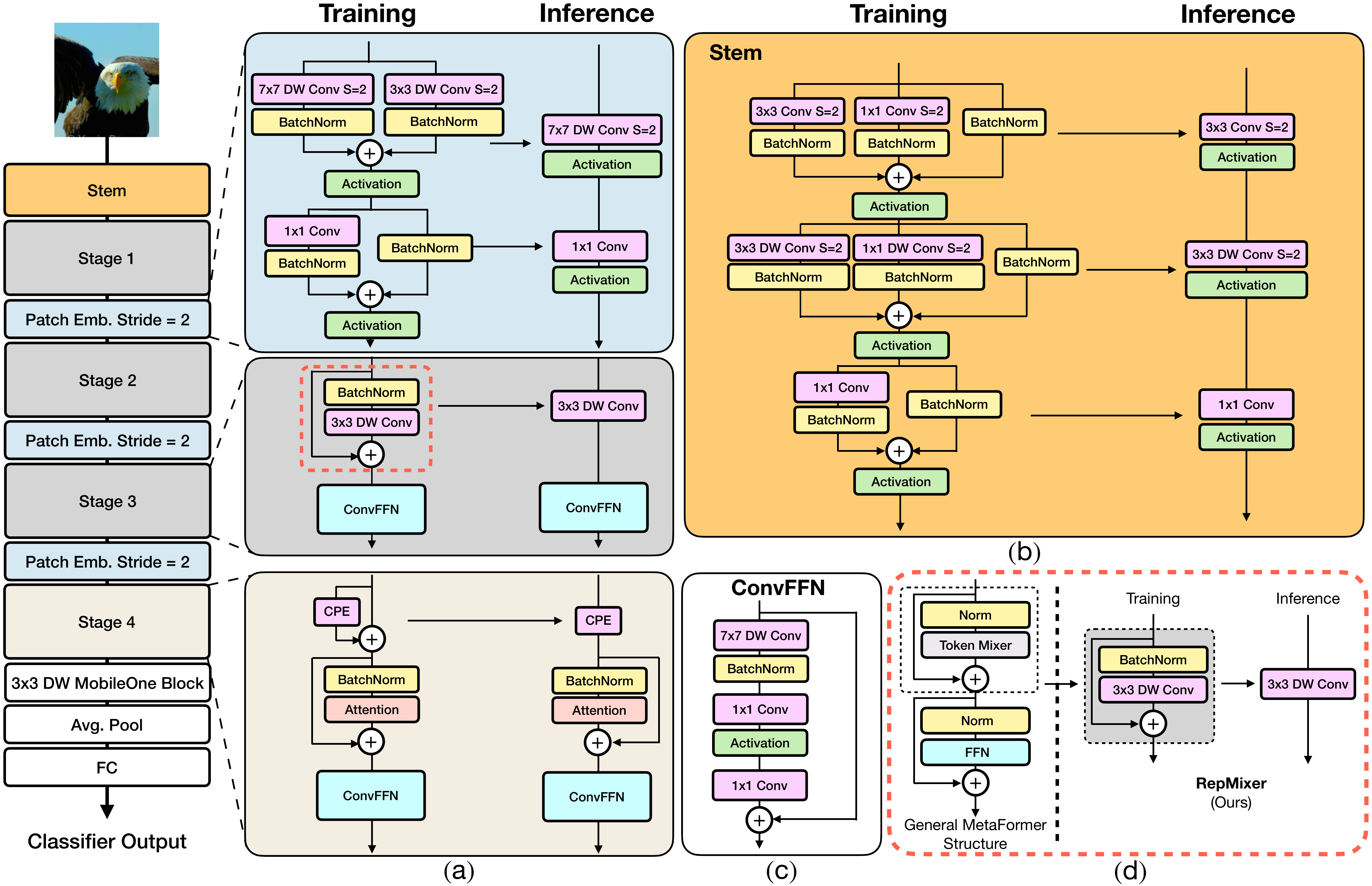}
    \caption{(a) Overview of FastViT architecture which decouples train-time and inference-time architecture. Stages 1, 2, and 3 have the same architecture and uses RepMixer for token mixing. In stage 4, self attention layers are used for token mixing. (b) Architecture of the convolutional stem. (c) Architecture of convolutional-FFN (d) Overview of \textit{RepMixer} block, which reparameterizes a skip connection at inference. }
    \label{fig:arch}
    \vspace{-4mm}
\end{figure*}

\begin{figure}[t]
    \centering
    \includegraphics[width=1.0\linewidth]{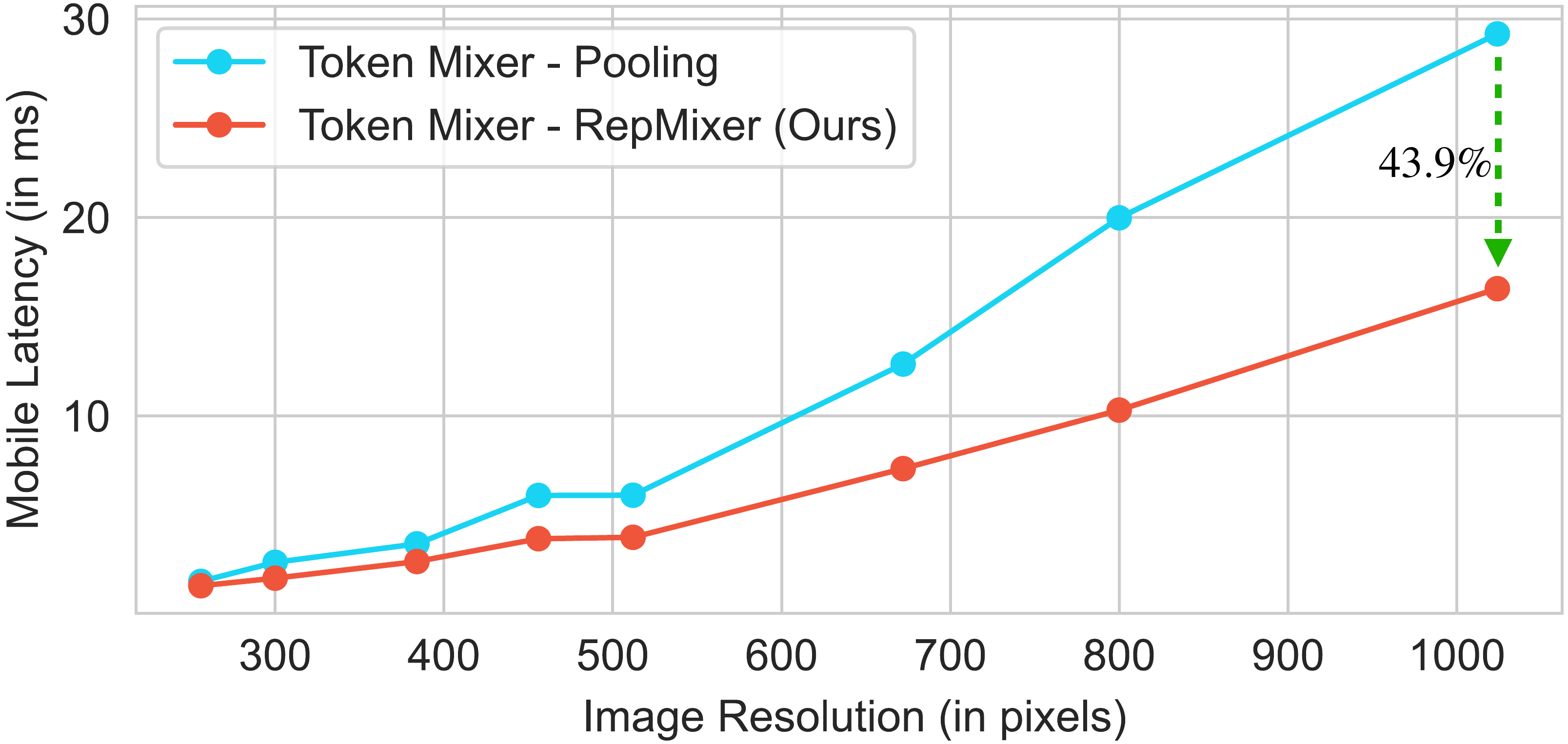}
    \caption{  
    Latency comparison of a MetaFormer (S12) architecture with Pooling and RepMixer as a choice for token mixing; measured on iPhone 12 Pro for various image resolutions. Both models have $\sim$1.8G FLOPs. Absence of a skip connection in RepMixer lowers the overall memory access cost leading to lower latency. }
    \label{fig:benefits_reparam}
    \vspace{-0.5cm}
\end{figure}

\subsection{Overview} 
\label{sec:overview}
FastViT is a hybrid transformer and has four distinct stages which operate at different scales as shown in Figure~\ref{fig:arch}. We detail all the FastViT variants in Table~\ref{tab:model_arch}.

FastViT uses \textit{RepMixer}, a token mixer that reparameterizes a skip connection, which helps in alleviating memory access cost (see Figure~\ref{fig:arch}d).
To further improve efficiency and performance, we replace dense k$\times$k convolutions commonly found in stem and patch embedding layers with its factorized version that uses train-time overparameterization (see Figure~\ref{fig:arch}a).

Self-attention~\cite{ViT} token mixers are computationally expensive, especially at higher resolutions~\cite{litv1, marin2021token}. While efficient versions of self-attention layers are explored in~\cite{Cmt, litv2}, we use large kernel convolutions as an efficient alternative to improve receptive field in early stages of the network architecture (see Figure~\ref{fig:arch}c). 

We analyze various architectural choices made in designing FastViT from a PoolFormer~\cite{metaformer} baseline in Table~\ref{tab:ablation_arch_choices} and detail our approach below.

\begin{table}
    \centering
    \scalebox{0.80}{
    \begin{tabular}{l@{\hspace{0.5\tabcolsep}}c@{\hspace{0.5\tabcolsep}}c@{\hspace{0.5\tabcolsep}}c@{\hspace{0.5\tabcolsep}}c@{\hspace{0.5\tabcolsep}}}
    \toprule
    \multirow{2}{*}{Architectural Choices} & Params & FLOPs & Mobile & Top-1 \\
     & (M) & (G) & Latency (ms) & (\%) \\ 
     \midrule
    \multicolumn{1}{l}{PoolFormer-S12 (Baseline)} & 11.9 & 1.8 & 1.50 & 77.2  \\
    \multicolumn{1}{l}{\quad + 224 $\rightarrow$ 256} & 11.9 & 2.4 & 2.25 & 77.6 \\
    \midrule
    Section \ref{sec:reparam_skip} \\
    \multicolumn{1}{l}{\quad + Pooling $\rightarrow$ RepMixer} & 11.9 & 2.4 & 1.58 & 78.5  \\
    \midrule
    Section \ref{sec:train_overparam} \\
    \multicolumn{1}{l}{\quad + Factorized dense conv.} & 8.7 & 1.7 & 1.26 & 78.0 \\
    \multicolumn{1}{l}{\quad + Train-Time Overparam.} & 8.7 & 1.7 & 1.26 & 78.9 \\
    \midrule
    Section \ref{sec:convnext_ffn} \\
    \multicolumn{1}{l}{\quad + LK. conv. FFN } & 8.7 & 1.8 & 1.33 & 79.4 \\
    \multicolumn{1}{l}{\quad + LK. conv. Patch Emb. } & 8.8 & 1.8 & 1.40 & 79.8 \\
    \bottomrule
    \end{tabular}
    
    }
        \caption{Analysis of architectural choices made to obtain FastViT-S12 variant, starting from PoolFormer-S12. ``LK." stands for Large Kernel.}
    \label{tab:ablation_arch_choices}
\end{table}

\begin{table}
\footnotesize
\centering
\setlength{\tabcolsep}{0.8pt}
\input{tables/models}
\caption{\textbf{Architecture details of FastViT variants.} Models with smaller embedding dimensions, i.e. [64, 128, 256, 512] are prefixed with ``S" and models that contain Self-Attention layers are prefixed with ``SA". Models with bigger embedding dimensions, i.e. [76, 152, 304, 608] are prefixed with ``M". Models with MLP expansion ratio less than 4 are prefixed with ``T". The number in the notation denotes total number of FastViT blocks. FLOP count is computed by \texttt{fvcore}\cite{fvcore} library.
}
\label{tab:model_arch}
\vspace{-4mm}
\end{table}

\subsection{FastViT}\label{sec:FastViT_arch}

\subsubsection{Reparameterizing Skip Connections}\label{sec:reparam_skip}

\paragraph{RepMixer} 
Convolutional mixing was first introduced in ConvMixer\cite{trockman2022convmixer}. For an input tensor $X$, the mixing block in the layer was implemented as,
\begin{equation}
Y = \texttt{BN(}\sigma\texttt{(DWConv(}X\texttt{)))} + X
\end{equation}
where $\sigma$ is a non-linear activation function and \texttt{BN} is Batch Normalization~\cite{Batch_Norm} layer and \texttt{DWConv} is depthwise convolutional layer. While this block was shown to be effective, in \textit{RepMixer}, we simply rearrange the operations and remove the non-linear activation function as shown below,
\begin{equation}
Y = \texttt{DWConv(BN(}X\texttt{)}+ X
\end{equation}\label{equation:repmix_train}
The main benefit of our design is that it can be reparameterized at inference time to a single depthwise convolutional layer as shown below and in Figure~\ref{fig:arch}d.
\begin{equation}
Y = \texttt{DWConv(}X\texttt{)} 
\end{equation}\label{equation:repmix_infer}

\vspace{-5mm}
\paragraph{Positional Encodings} 
We use conditional positional encodings~\cite{Twins, CPE} that is dynamically generated and conditioned on the local neighborhood of the input tokens. These encodings are generated as a result of a depth-wise convolution operator and are added to the patch embeddings. Note the lack of non-linearities in this group of operations, hence this block is reparameterized as shown in Figure~\ref{fig:arch}a.

\paragraph{Empirical Analysis} In order to verify the benefits of reparameterizing skip connections, we ablate over the using one of the most efficient (in terms of latency) token mixer, i.e. Pooling and RepMixer in a MetaFormer S12 architecture. Both the models being ablated have $\sim$1.8G FLOPs. We time the models for various input resolutions starting from 224$\times$224 to 1024$\times$1024 on an iPhone 12 Pro mobile device. From Figure~\ref{fig:benefits_reparam}, we see that RepMixer significantly improves over Pooling, especially at higher resolutions. At 384$\times$384, using RepMixer will lower the latency by 25.1\% and at larger resolutions like 1024$\times$1024, latency is lowered significantly by 43.9\%.

\vspace{-3.5mm}
\subsubsection{Linear Train-time Overparameterization}\label{sec:train_overparam}

In order to further improve efficiency (parameter count, FLOPs, and latency), we replace all dense k$\times$k convolutions with its factorized version, i.e. k$\times$k depthwise followed by 1$\times$1 pointwise convolutions. However, the lower parameter count from factorization can diminish the capacity of the model. In order to increase capacity of the factorized layers, we perform linear train-time overparameterization as described in MobileOne~\cite{mobileone}.
MobileOne-style overparameterization in stem, patch embedding, and projection layers help in boosting performance. From Table~\ref{tab:ttop_ablation}, we note that train-time overparameterization improves Top-1 accuracy on ImageNet by 0.6\% on FastViT-SA12 model. On a smaller FastViT-S12 variant, Top-1 accuracy improves by 0.9\% as shown in Table~\ref{tab:ablation_arch_choices}.

However, train-time overparameterization results in increased training time due to computational overhead from the added branches. In our architecture, we only overparameterize those layers that replace dense k$\times$k convolutions with its factorized form as described above. These layers are found in the convolutional stem, patch embedding and projection layers. The computational cost incurred in these layers are lower than the rest of the network, hence overparameterizing these layers do not result in significant increases to train time. For example, FastViT-SA12 takes 6.7\% longer and FastViT-SA36 takes 4.4\% longer to train with train-time overparameterization as opposed to training those variants without it under the same settings described in Section~\ref{sec:imagenet_exps}.

\vspace{-3.5mm}
\subsubsection{Large Kernel Convolutions}\label{sec:convnext_ffn}
The receptive field of RepMixer is local compared to self-attention token mixers. However, self-attention based token mixers are computationally expensive. A computationally efficient approach to improve the receptive field of early stages that do not use self-attention is by incorporating depthwise large kernel convolutions. We introduce depthwise large kernel convolutions in FFN and patch embedding layers. 
From Table~\ref{tab:large_kernel_ablations}, we note that variants using depthwise large kernel convolutions can be highly competitive to variants using self-attention layers while incurring a modest increase in latency. When we compare V5 with V3, model size increases by 11.2\%, and latency increases by a factor of 2.3$\times$ for a relatively small gain of 0.4\% in Top-1 accuracy. V2 is larger than V4 by 20\% and has 7.1\% higher latency than V4 while attaining similar Top-1 accuracy on ImageNet. Further ablations on kernel sizes and latency is provided in the supplementary materials. In Table~\ref{tab:ablation_arch_choices}, we ablate over large kernel convolutions in FFN and patch embedding layers. Overall, large kernel convolutions provide 0.9\% improvement in Top-1 accuracy on FastViT-S12. 

The architecture of our FFN and patch embedding layer is shown in Figure~\ref{fig:arch}. Our FFN block has a structure similar to ConvNeXt~\cite{ConvNext} block with a few key differences, see Figure~\ref{fig:arch}c. We use Batch Normalization as opposed to Layer Normalization, as it can be fused with the preceding layer at inference. Also, it does not require additional reshape operations to obtain appropriate tensor layout for LayerNorm as done in the original implementation of ConvNeXt block. 

Along with increased receptive field, large kernel convolutions help in improving model robustness as observed in~\cite{wang2022robustcnn} and convolutional-FFN blocks generally tend to be more robust than vanilla-FFN blocks as observed in~\cite{mao2022robust}. Hence, incorporating large kernel convolutions is an efficient way to improve model performance and robustness. 

\begin{table}
    \centering
    \scalebox{0.68}{
    \begin{tabular}{l|c|c|c}
    \toprule
    Model   & Ablation  & Top-1 (\%) & Train Time (hrs)  \\ 
    \midrule
    \multirow{2}{*}{FastViT-SA12} & Without Train-Time Overparam. & 80.0 & 31.3 \\
                                & With Train-Time Overparam. & \textbf{80.6} & 33.4 \\
    \midrule                                
    \multirow{2}{*}{FastViT-SA36} & Without Train-Time Overparam. & 83.3 & 73.5 \\
                                & With Train-Time Overparam. & \textbf{83.6} & 76.7 \\                                
    \bottomrule
    \end{tabular}
    
    }
        \caption{Comparison of FastViT variants with and without linear train-time overparameterization when trained on ImageNet-1k dataset. Train time is wall clock time elapsed at the end of a training run.}
    \label{tab:ttop_ablation}
    \vspace{-0.4cm}
\end{table}

\begin{table}
    \centering
    \scalebox{0.78}{
    \begin{tabular}{c|c|c|c|c|c|c|c}
    \toprule
    \multirow{2}{*}{Variant} & \multicolumn{4}{c|}{Stages}  & Params & Top-1 & Mobile \\
    \cmidrule{2-5}
    & 1 & 2 & 3 & 4 & (M) & (\%) & Latency (ms) \\ 
    \midrule
    \multicolumn{8}{c}{Standard Setting} \\
    \midrule
    V1 & RM & RM & RM & RM & 8.7  & 78.9 & 1.3 \\
    V2 & RM & RM & RM & SA & 10.8 & 79.9 & 1.5 \\
    V3 & RM & RM & SA & SA & 12.4 & 81.0 & 3.7 \\
    \midrule
    \multicolumn{8}{c}{Large Kernel Convolutions (7$\times$7)} \\
    \midrule
    V4 & RM & RM & RM & RM & 8.8  & 79.8  & 1.4 \\
    V5 & RM & RM & RM & SA & 10.9 & 80.6  & 1.6 \\
    \bottomrule
    \end{tabular}
    
    }
        \caption{Ablation on using large kernel convolutions as a substitute for self-attention layers. ``RM" indicates [RepMixer-FFN] block is used in the stage. ``SA" indicates [Self Attention-FFN] block is used in the stage. Standard setting uses 3x3 factorized convolutions in patch embedding and stem layers and 1$\times$1 convolutions for FFN. In variants V4 and V5, large kernel convolutions (7$\times$7) are used in patch embedding and FFN layers.}
    \label{tab:large_kernel_ablations}
\end{table}

%% file: tables/models.tex
\newcommand{\modelname}{FastViT}
\newcommand{\tabincell}[2]{\begin{tabular}{@{}#1@{}}#2\end{tabular}}

\newcommand{\blockc}[4]{
$\begin{bmatrix}
	\begin{array}{l}
	R_1=#1 \\
	N_1=#2 \\
	E_1=#3 \\
	\end{array}
\end{bmatrix} \times #4$
}

\newcommand{\sblock}[3]{
$\begin{matrix}
E_{#1}=#2 \\
L_{#1}=#3 \\
\end{matrix}$
}

\newcommand{\poollayer}{
Mixer & \multicolumn{7}{c}{RepMixer}\\
\cline{4-11}
}

\newcommand{\stitle}[7]{
\multirow{5}{*}{#1} & \multirow{5}{*}{\scalebox{1}{$\frac{H}{#2}\times \frac{W}{#2}$}} & \multicolumn{2}{c|}{\multirow{2}{*}{\tabincell{c}{Patch Embed.}}} & \multicolumn{7}{c}{$7 \times 7$ MobileOne Style, stride $#4$} \\
\cline{5-11}
    &    &    \multicolumn{2}{c|}{} & $#5$ & \multicolumn{5}{c|}{$#6$} & $#7$ \\
\cline{3-11}
& & \multirow{3}{*}{\tabincell{c}{\modelname{}\\Block}} 
}

\begingroup
\renewcommand{\arraystretch}{1.1}
\begin{tabular}{c|c|c|c|c|c|c|c|c|c|c}
\toprule
  \multirow{2}{*}{Stage} & \multirow{2}{*}{\#Tokens} & \multicolumn{2}{c|}{\multirow{2}{*}{Layer Spec.}} & \multicolumn{7}{c}{\modelname{}} \\
\cline{5-11}
 & & \multicolumn{2}{c|}{} & T8 & T12 & S12 & SA12 & SA24 & SA36 & MA36 \\
\cline{1-11}
\multirow{3}{*}{Stem} & \multirow{3}{*}{\scalebox{0.8}{$H \times W$}} & \multicolumn{2}{c|}{\multirow{3}{*}{Conv.}} & \multicolumn{7}{c}{$3 \times 3$, stride $2$} \\
\cline{5-11}
& & \multicolumn{2}{c|}{} & \multicolumn{7}{c}{$3 \times 3$ MobileOne Style, stride $2$} \\
\cline{5-11}
& & \multicolumn{2}{c|}{} & 48 & \multicolumn{5}{c|}{64} & 76 \\
\cline{1-11}
\stitle{1}{4}{7}{2}{48}{64}{76}    & \poollayer
 & & & Exp. & \multicolumn{2}{c|}{3} & \multicolumn{5}{c}{4} \\
\cline{4-11}
 & & & Blocks & 2 & 2 & 2 & 2 & 4 & 6 & 6 \\
\hline
\stitle{2}{8}{3}{2}{96}{128}{152}  & \poollayer
 & & & Exp. & \multicolumn{2}{c|}{3} & \multicolumn{5}{c}{4} \\
 \cline{4-11}
 & & & Blocks & 2 & 2 & 2 & 2 & 4 & 6 & 6 \\
\hline
\stitle{3}{16}{3}{2}{192}{256}{304} & \poollayer
 & & & Exp. & \multicolumn{2}{c|}{3} & \multicolumn{5}{c}{4} \\
\cline{4-11}
 & & & Blocks & 4 & 6 & 6 & 6 & 12 & 18 & 18 \\
\hline
\stitle{4}{32}{3}{2}{384}{512}{608} & Mixer & \multicolumn{3}{c|}{RepMixer} & \multicolumn{4}{c}{Attention}\\
\cline{4-11}
 & & & Exp. & \multicolumn{2}{c|}{3} & \multicolumn{5}{c}{4} \\
 \cline{4-11}
 & & & Blocks & 2 & 2 & 2 & 2 & 4 & 6 & 6 \\
\hline
\multicolumn{4}{c|}{Parameters~(M)} & 3.6 & 6.8 & 8.8 & 10.9 & 20.6 & 30.4 & 42.7 \\
\hline
\multicolumn{4}{c|}{FLOPs~(G)} & 0.7 & 1.4 & 1.8 & 1.9 & 3.8 & 5.6 & 7.9 \\
\bottomrule
\end{tabular}
\endgroup

%% file: experiments.tex
\section{Experiments}
\label{sec:experiments}

\subsection{Image Classification}\label{sec:imagenet_exps}

We report results on the the widely used ImageNet-1K dataset~\cite{ImageNet-1K} which contains $\sim$1.3M training images and 50K validation images. We follow the training recipe prescribed in ~\cite{deit, metaformer}, i.e. the models are trained for 300 epochs using AdamW optimizer with weight decay of 0.05 and peak learning rate $10^{-3}$ for a total batch size of 1024. The number of warmup epochs is set to 5 and cosine schedule is used to decay the learning rate. Our implementation uses timm library~\cite{timm} and all the models were trained on 8 NVIDIA A100 GPUs. See supplementary materials for details on hyper parameters used for all the variants. For 384$\times$384 input size, we fine-tune the models for 30 epochs with weight decay of $10^{-8}$ and learning rate of $5\times10^{-5}$ and batch size of 512 following~\cite{ConvNext}. To measure latency, we use the input sizes corresponding to the respective methods. For iPhone latency measurements, we export the models using Core ML Tools (v6.0) and run it on iPhone12 Pro Max with iOS 16 and batch size is set to 1 for all the models. We follow the same protocol as~\cite{mobileone}. For GPU latency measurements, we export the traced model to TensorRT (v8.0.1.6) format and run it on NVIDIA RTX-2080Ti with batch size of 8. We report the median latency estimate from 100 runs.

\begin{table}[]
    \centering

    \scalebox{0.72}{
    \begin{tabular}{l| @{\hspace{0.5\tabcolsep}} c @{\hspace{1.0\tabcolsep}} c @{\hspace{1.0\tabcolsep}}c @{\hspace{1.0\tabcolsep}}c @{\hspace{1.0\tabcolsep}}c| @{\hspace{0.5\tabcolsep}}c}
    \toprule
    \multirow{2}{*}{Model}        & Eval    & Param     & FLOPs    & GPU  & Mobile       & Top-1  \\ 
    % \cmidrule{5-6}
                                    & Image &        &       & Latency     & Latency  &  Acc.      \\   
                    & Size & (M) & (G) & (ms)  & (ms) & (\%) \\
                                    
                                    \midrule

    MobileOne-S0\cite{mobileone} & 224 & 2.1 & 0.3 & \textbf{1.0} & \textbf{0.8} & 71.4 \\ 
    \textbf{FastViT-T8} & 256 & 3.6 & 0.7 & 1.7 & \textbf{0.8} & \textbf{75.6} \\ 
    \midrule
    MobileOne-S3\cite{mobileone} & 224 & 10.1 & 1.8 & \textbf{2.1} & 1.5 & 78.1 \\ 
    
    CMT-$\text{T}^*$\cite{Cmt}  & 160 & 9.5 & 0.6 & 4.5 & 3.8 & \textbf{79.1} \\
    EfficientNet-B1\cite{EfficientNet} & 256 & 7.8 & 0.7 & 3.2 & 2.5 & \textbf{79.1} \\
    \textbf{FastViT-T12} & 256 & 6.8 & 1.4 & \textbf{2.1} & \textbf{1.2} & \textbf{79.1} \\

    \midrule
    MobileOne-S4\cite{mobileone} & 224 & 14.8 & 2.9 & 2.6 & 1.9 & 79.4 \\
    RSB-ResNet50\cite{rsbresnets} & 224 & 25.6 & 4.1 & 2.3 & 2.6 & \textbf{79.8} \\
    DeiT-S\cite{deit}  & 224 & 22.0 & 4.6 & 5.2 & 5.3 & \textbf{79.8} \\
    \textbf{FastViT-S12} & 256 & 8.8 & 1.8 & \textbf{2.2} & \textbf{1.4} & \textbf{79.8} \\
    
    \midrule
    
    RegNetY-8G\cite{RegNet} & 224 & 39.2 & 8.0 & 5.8 & 3.8 & 79.9 \\
    EfficientNet-B2\cite{EfficientNet} & 288 & 9.2 & 1.0  & 4.1 & 3.5 & 80.1 \\
    PoolFormer-S24\cite{metaformer} & 224  & 21.0 & 3.4 & 15.7 & 2.5 & 80.3 \\
    RegNetY-16G \cite{RegNet}  & 224 & 83.6 & 16.0 & 8.1 & 7.5 & 80.4 \\
    \textbf{FastViT-SA12} & 256 & 10.9 & 1.9 & \textbf{2.5} & \textbf{1.6} & \textbf{80.6} \\
    
    \midrule
    
    ResNeSt50\cite{zhang2022resnest} & 224 & 27.5 & 5.4 & 51.5 & 29.9 & 81.1 \\
    Swin-T\cite{Swin} & 224 & 29.0 & 4.5 & -  & 9.3 & 81.3\\
    PoolFormer-S36\cite{metaformer}  & 224 & 31.0 & 5.0 & 23.7 & 3.5 & 81.4 \\
    EfficientNet-B3\cite{EfficientNet} & 300 & 12.0 & 1.8 & 5.3 & 7.1 & 81.6 \\
    CvT-13\cite{cvt} & 224 & 20.1 & 4.5 & 18.1 & 62.6 & 81.6 \\
    CoAtNet-0\cite{CoAtNet} & 224 & 25.0 & 4.2 & - & 6.7 & 81.6 \\
    CMT-$\text{XS}^*$\cite{Cmt} & 192 & 15.2 & 1.5 & 8.3 & 6.3 & 81.8 \\
    DeiT-B \cite{deit} & 224 & 86.0 & 17.5 & 11.6 & 11.6 & 81.8 \\
    RSB-ResNet152\cite{rsbresnets} & 224 & 60.2 & 11.6 & 5.4 & 4.7 & 81.8 \\
	LITv2-S\cite{litv2} & 224 & 28.0 & 3.7 & - & - & 82.0 \\
	ConvNeXt-T\textsuperscript{\textdagger}\cite{ConvNext} & 224 & 29.0 & 4.5 & 6.3 & 3.7 & 82.1 \\ 
    PoolFormer-M48\cite{metaformer} & 224 & 73.0 & 11.6 & 30.2 & 7.6 & 82.5 \\
    \textbf{FastViT-SA24} & 256 & 20.6 & 3.8 & \textbf{3.8} & \textbf{2.6} & \textbf{82.6} \\
    \midrule
        
    ResNeSt101\cite{zhang2022resnest} & 224 & 48.2 & 10.2 & 75.5 & 37.7 & 83.0 \\
    Swin-S\cite{Swin}   & 224  & 50.0  & 8.7 & -  & 13.0 & 83.0        \\
    ConvNeXt-S\textsuperscript{\textdagger}\cite{ConvNext}  & 224  & 50.0 & 8.7 & 10.3 & 5.4 & 83.1 \\
    MOAT-0\cite{moat} & 224 & 27.8 & 5.7 & - & 4.9 & 83.3 \\
    CoAtNet-1\cite{CoAtNet} & 224 & 42.0 & 8.4 & - & 13.4 & 83.3 \\
    Swin-B \cite{Swin}  & 224 & 88.0 & 15.4 & - & 18.5 & 83.5  \\
    CMT-$\text{S}^*$\cite{Cmt} & 224 & 25.1 & 4.0 & 13.8 & 15.6 & 83.5 \\
    MaxViT-T\cite{maxvit} & 224 & 31.0 & 5.6 & - & 31.5 & \textbf{83.6} \\
    LITv2-B\cite{litv2} & 224 & 87.0 & 13.2 & - & - & \textbf{83.6} \\
    EfficientNet-B5\cite{EfficientNet}  & 456 & 30.0 & 9.9 & 24.2  & 22.1  & \textbf{83.6}  \\
    NFNet-F0\cite{nfnets} & 256 & 71.5 & 12.4 & 6.2 & 6.4 & \textbf{83.6} \\
    \textbf{FastViT-SA36} & 256 & 30.4 & 5.6 & \textbf{5.2} & \textbf{3.5} & \textbf{83.6} \\
    \midrule

    ConvNeXt-B\textsuperscript{\textdagger}\cite{ConvNext}             &224        & 89.0      & 15.4       &  13.6     & 8.4    & 83.8\\
    CoAtNet-0 & 384 & 25.0 & 13.4 & - & 20.1 & \textbf{83.9} \\
    EfficientNetV2-S\cite{efficientnet_v2_quoc} & 384 & 22.0 & 8.8 & 8.2 & 7.1 & \textbf{83.9}\\
     \textbf{FastViT-MA36} & 256 & 42.7 & 7.9 & \textbf{6.7} & \textbf{4.5} & \textbf{83.9} \\

    \midrule

    EfficientNet-B6\cite{EfficientNet}  & 528 & 43.0 & 19.0 & 44.3 & 51.7  & 84.0  \\
    CoAtNet-2\cite{CoAtNet} & 224 & 75.0 & 15.7 & - & 19.6 & 84.1 \\
    MOAT-1\cite{moat} & 224 & 41.6 & 9.1 & - & 8.3 & 84.2 \\
    EfficientNet-B7\cite{EfficientNet}  & 600 & 66.0 & 37.0 & 72.6 & 85.5 & 84.3  \\
    CMT-$\text{B}^*$\cite{Cmt} & 256 & 45.7 & 9.3 & 26.9 & 20.6 & \textbf{84.5} \\
    Swin-B\cite{Swin}   & 384  & 88.0  & 47.0 & -  & 57.2 & \textbf{84.5}        \\
    MaxViT-S\cite{maxvit} & 224 & 69.0 & 11.7 & - & 48.7 & \textbf{84.5} \\
    \textbf{FastViT-SA36} & 384 & 30.4 & 12.6 & \textbf{12.8} & \textbf{7.4} & \textbf{84.5} \\
    
    \midrule

    MOAT-0\cite{moat} & 384 & 27.8 & 18.2 & - & 24.3 & 84.6 \\
    NFNet-F1\cite{nfnets} & 320 & 132.6 & 35.5 & 16.9 & 16.1 & 84.7 \\
    MOAT-2\cite{moat} & 224 & 73.4 & 17.2 & - & 12.6 & 84.7 \\
    LITv2-B\cite{litv2} & 384 & 87.0 & 39.7 & - & - & 84.7 \\
    \textbf{FastViT-MA36} & 384 & 42.7 & 17.7 & \textbf{16.5} & \textbf{9.2} & \textbf{84.9} \\
    
    \bottomrule
    \end{tabular}
    
    }
        \caption{Comparison of different state-of-the-art methods on ImageNet-1k classification. HardSwish is not well supported by Core ML, $*$ denotes we replace it with GELU for fair comparison. ``$\dagger$" denotes that model has been modified from original implementation for efficient deployment.  
        Models which could not be reliably exported either by TensorRT or Core ML Tools are annotated by ``-".}
    \label{tab:ImageNet_1K}
\end{table}

\hfill \break
\textbf{Comparison with SOTA Models}
In Table~\ref{tab:ImageNet_1K}, we compare our models against recent state-of-the-art models on ImageNet-1k dataset. 
For a fair comparison, we modify ConvNeXt~\cite{ConvNext} from official implementation by avoiding costly reshape operations, see supplementary materials for more details. 
We were unable to reliably export LITv2~\cite{litv2} due to poor support for deformable convolutions in either library. Our model obtains the best accuracy-latency trade-off when compared to recent state-of-the-art models on two different compute fabrics, i.e. desktop-grade GPU and mobile device. Our models improve over LITv2~\cite{litv2} on both parameter count and FLOPs, at Top-1 accuracy of 84.9\%, FastViT-MA36 is 49.3\% smaller and consumes 55.4\% less FLOPs than LITv2-B.  FastViT-S12 is 26.3\% faster than MobileOne-S4~\cite{mobileone} on iPhone 12 Pro and 26.9\% faster on GPU. At Top-1 accuracy of 83.9\%, FastViT-MA36 is 1.9$\times$ faster than an optimized ConvNeXt-B model on iPhone 12 Pro and 2.0$\times$ faster on GPU. At Top-1 accuracy of 84.9\%, FastViT-MA36 is just as fast as NFNet-F1~\cite{nfnets} on GPU while being 66.7\% smaller and using 50.1\% less FLOPs and 42.8\% faster on mobile device.  

\hfill \break
\textbf{Knowledge distillation}
We report performance of our models when trained with distillation objective in Table~\ref{tab:ImageNet_1K_distill}. We follow the setting described in DeiT~\cite{deit}, with RegNet-16GF~\cite{RegNet} as the teacher model. Following DeiT~\cite{deit}, we use hard distillation where hard decision of the teacher is set as true label. Our models are trained for 300 epochs. Unlike~\cite{deit, cait, li2022efficientformer}, we do not introduce an additional classification head for distillation. FastViT outperforms recent state-of-the-art model EfficientFormer~\cite{li2022efficientformer}. FastViT-SA24 attains similar performance as EfficientFormer-L7 while having 3.8$\times$ less parameters, 2.7$\times$ less FLOPs and 2.7$\times$ lower latency.

\begin{table}
    \centering
    \scalebox{0.74}{
    \begin{tabular}{l| @{\hspace{0.5\tabcolsep}} c @{\hspace{1.0\tabcolsep}} c @{\hspace{1.0\tabcolsep}}c @{\hspace{1.0\tabcolsep}}c @{\hspace{1.0\tabcolsep}}c| @{\hspace{0.5\tabcolsep}}c}
    \toprule
    \multirow{2}{*}{Model}        & Eval    & Param     & FLOPs    & GPU  & Mobile       & Top-1  \\ 
    
                                    & Image &        &       & Latency     & Latency  &  Acc.      \\   
                    & Size & (M) & (G) & (ms)  & (ms) & (\%)      \\
                    \midrule
    % \midrule

    \textbf{FastViT-T8} & 256 & 3.6 & 0.7 & 1.7 & \textbf{0.8} & \textbf{76.7} \\
    \midrule
    \textbf{FastViT-T12} & 256 & 6.8 & 1.4 & 2.1 & \textbf{1.2} & \textbf{80.3} \\
    \midrule
    
    CaiT XXS-36\cite{cait} & 224 & 17.3 & 3.8 & 15.8 & - & 79.7 \\
    \textbf{FastViT-S12} & 256 & 8.8 & 1.8 & \textbf{2.2} & \textbf{1.4} & \textbf{80.9} \\
    \midrule
    
    DeiT-S\cite{deit} & 224 & 22 & 4.6 & 5.2 & 6.7 & 81.2 \\
    \textbf{FastViT-SA12}  & 256 & 10.9 & 1.9 & \textbf{2.5}  & \textbf{1.6} & \textbf{81.9} \\
    
    \midrule
    
    EfficientFormer-L3\cite{li2022efficientformer}  & 224 & 31.3 & 3.9 & \textbf{3.6} & 3.0 & 82.4 \\
    EfficientFormer-L7\cite{li2022efficientformer}  & 224 & 82.1 & 10.2 & 7.5 & 7.0 & 83.3 \\
    DeiT-B\cite{deit} & 224 & 87 & 17.6 & 11.6 & 11.6 & \textbf{83.4} \\
    \textbf{FastViT-SA24}  & 256 & 20.6 & 3.8 & 3.8 & \textbf{2.6} & \textbf{83.4} \\
    
    \midrule
    
    CaiT S-24\cite{cait} & 224 & 46.9 & 9.4 & 19.1 & - & 83.5 \\
    CaiT XS-24\cite{cait} & 384 & 26.6 & 19.3 & 77.7 & - & 84.1 \\
    \textbf{FastViT-SA36}  & 256 & 30.4 & 5.6 & \textbf{5.2} & \textbf{3.5} & \textbf{84.2} \\
    \midrule
    
    \textbf{FastViT-MA36}  & 256 & 42.7 & 7.9 & \textbf{6.7} & \textbf{4.5} & \textbf{84.6} \\
    
    \bottomrule
    \end{tabular}
    
    }
        \caption{Comparison of different state-of-the-art methods on ImageNet-1k classification when trained using distillation objective specified by~\cite{deit}. Models which could not be reliably exported either by TensorRT or Core ML Tools are annotated by ``-".}
    \label{tab:ImageNet_1K_distill}
    \vspace{-.2cm}
\end{table}

\subsection{Robustness Evaluation}
We evaluate our models for out-of-distribution robustness on the following benchmarks -- (i) ImageNet-A~\cite{imageneta}, a dataset that contains naturally occurring examples that are misclassified by ResNets; (ii) ImageNet-R~\cite{imagenetr}, a dataset that contains natural renditions of ImageNet object classes with different textures and local image statistics; (iii) ImageNet-Sketch~\cite{imagenetsketch}, a dataset that contains black and white sketches of all ImageNet classes, obtained using google image queries; and (iv) ImageNet-C~\cite{imagenetc}, a dataset that consists of algorithmically generated corruptions (blur, noise) applied to the ImageNet test-set. We evaluate our models using the implementation provided by~\cite{mao2022robust}. Our models are more robust than recent vision transformers while being faster than pure-CNN based models which exhibit low robustness as seen Table~\ref{tab:robustness}. Architectural choices like using a large kernel convolutions in FFN and patch-embedding layers in combination with self-attention layers helps in improving the robustness of our model as discussed in Section~\ref{sec:convnext_ffn}. All the models compared in Table~\ref{tab:robustness} are trained only on ImageNet-1k dataset using similar training recipes. From Table~\ref{tab:robustness}, our model is highly competitive to RVT and ConvNeXt, in fact FastViT-M36 has better clean accuracy, better robustness to corruptions and similar out-of-distribution robustness as ConvNeXt-S which has 6.1M more parameters and has 10\% more FLOPs than our model.

\begin{table}[t]
\centering
\scalebox{0.73}{
\begin{tabular}{l@{\hspace{0.95\tabcolsep}}c@{\hspace{0.95\tabcolsep}}c@{\hspace{0.95\tabcolsep}}c@{\hspace{0.95\tabcolsep}}c@{\hspace{0.95\tabcolsep}}c@{\hspace{0.95\tabcolsep}}c@{\hspace{0.95\tabcolsep}}c}
\toprule
Model   & \#Params & \#FLOPs & Clean & IN-C ($\downarrow$) & IN-A & IN-R & IN-SK \\
\midrule
PVT-Tiny & 13.2 & 1.9 & 75.0 & 79.6 & 7.9 & 33.9 & 21.5 \\
RVT-Ti & 8.6 & 1.3 & \underline{78.4} & \textbf{58.2} & \underline{13.3} & \textbf{43.7} & \textbf{30.0} \\
\textbf{FastViT-SA12} & 10.9 & 1.9 &\textbf{80.6} & \underline{62.2} & \textbf{17.2} & \underline{42.6} & \underline{29.7} \\
\midrule
PVT-Small    & 24.5 & 3.8 & \underline{79.9} & 66.9 & \underline{18.0} & 40.1 & 27.2   \\
ResNet-50*  & 25.6 & 4.1 & 79.0 & 65.5 & 5.9 & \underline{42.5} & \underline{31.5}      \\
ResNeXt50-32x4d & 25.0 & 4.3 & 79.8 & \underline{64.7} & 10.7 & 41.5 & 29.3 \\
\textbf{FastViT-SA24}  & 20.6 & 3.8 & \textbf{82.6} & \textbf{55.3} & \textbf{26.0} & \textbf{46.5} & \textbf{34.0} \\
\midrule
Swin-T     & 28.3 & 4.5 & 81.2 & 62.0 & 21.6 & 41.3 & 29.1    \\ 
ConViT-S   & 27.8 & 5.4 & 81.5 & \textbf{49.8} & 24.5 & 45.4 & 33.1    \\
RVT-S & 22.1 & 4.7 & 81.7 & \underline{50.1} & 24.1 & 46.9 & 35.0  \\ 
ConvNeXt-T & 28.6 & 4.5 & 82.1 & 53.2 & 24.2 & \underline{47.2} & 33.8  \\ 
EfficientNet-B4 & 19.3 & 4.2 & \underline{83.0} & 71.1 & \underline{26.3} & 47.1 & \underline{34.1} \\
\textbf{FastViT-SA36} & 30.4 & 5.6 & \textbf{83.6} & 51.8 & \textbf{32.3} & \textbf{48.1} & \textbf{35.8} \\
\midrule
ConvNeXt-S & 50.0 & 8.7 & 83.1 & 51.2 & 31.2 & \textbf{49.5} & \textbf{37.1} \\ 
\textbf{FastViT-MA36} & 42.7 & 7.9 & \textbf{83.9} & \textbf{49.9} & \textbf{34.6} & \textbf{49.5} & 36.6  \\ 
\bottomrule

\end{tabular}
}
\caption{Results on robustness benchmark datasets. Models are grouped based on FLOPs. Performance of competing models is reported by~\cite{mao2022robust} and \cite{ConvNext}. *ResNet-50 model is trained with AugMix to improve robustness as reported in ~\cite{mao2022robust}. For ImageNet-C mean corruption error is reported (lower is better) and for other datasets Top-1 accuracy is reported (higher is better). The best results are in bold and second best results are underlined.}
\label{tab:robustness}
% \vspace{-0.3cm}
\end{table}

\subsection{3D Hand mesh estimation}
Recent works on real-time 3D hand mesh estimation introduce complex mesh regression layers over CNN based backbones. The backbones usually belong to ResNet or MobileNet family of architectures with the exception of METRO and MeshGraphormer which use HRNets~\cite{hrnets} for feature extraction. While most hardware devices are highly optimized for feature extraction from 2D CNNs, the same is not true for the complex mesh regression heads used in these methods. In our method, we replace the complex mesh regression head with a simple regression module which regresses weak perspective camera, pose and shape parameters of the MANO model~\cite{manomodel}. We argue that using a feature extraction backbone that learns a good representation for underlying images can alleviate learning challenges in mesh regression. While other real-time methods compensate for weak feature extraction backbone with complex mesh regression layers, we use a better feature extraction backbone with a simple mesh regression layer. We compare our approach with other published methods on the FreiHand~\cite{freihand} dataset. For a fair comparison, we cite results of methods that only used FreiHand dataset for training as some methods either pre-train, train, or fine-tune with additional pose datasets. We only use ImageNet-1k dataset for pre-training and then train exclusively on the FreiHand dataset using the experimental setup described in ~\cite{metro_hand}. For more details, please see supplementary materials. From Table~\ref{tab:freihand}, amongst real-time methods, our method outperforms other methods on all joint and vertex error related metrics while being 1.9$\times$ faster than MobileHand~\cite{mobilehand} and 2.8$\times$ faster than recent state-of-art MobRecon.

\begin{table}[t]
\centering
\scalebox{0.77}{
\begin{tabular}{l @{\hspace{0.5\tabcolsep}} c @{\hspace{0.5\tabcolsep}} | c @{\hspace{0.5\tabcolsep}} c @{\hspace{0.5\tabcolsep}} c @{\hspace{0.5\tabcolsep}} c @{\hspace{0.5\tabcolsep}} c}
\toprule
Method  &  Backbone  & PJ $\downarrow$ & PV $\downarrow$ & F@5 $\uparrow$ & F@15 $\uparrow$ & FPS $\uparrow$ \\
\midrule
\multicolumn{7}{c}{Non Real-Time methods}\\
\midrule
HIU-DMTL \cite{hiu_dtml}           & Customized & 7.1 & 7.3   & 0.699 & 0.974 & 5  \\
METRO \cite{metro_hand}  & HRNet & 6.3 & 6.5 & 0.731 & 0.984 & 4.2 \\
MeshGraphormer \cite{meshgraphormer}  & HRNet & 5.9  & 6.0 & 0.764 & 0.986 & 2.6 \\
\midrule
\multicolumn{7}{c}{Real-Time methods}\\
\midrule
Hasson \etal \cite{obman} & ResNet18 & - & 13.2 & 0.436 & 0.908 & 20 \\
MobileHand \cite{mobilehand}   & MobileNetV3 & - & 13.1 & 0.439 & 0.902 & 110  \\
YoutubeHand \cite{youtube_hand} & ResNet50  & 8.4  & 8.6  & 0.614 & 0.966 & 60  \\
I2L-MeshNet \cite{i2l_meshnet}        & ResNet50 & 7.4 & 7.6   & 0.681 & 0.973 & 33.3 \\
Pose2Mesh \cite{pose2mesh}    & Customized & 7.4 & 7.6   & 0.683 & 0.973 & 20  \\
I2UV-HandNet \cite{i2uv_handnet}      & ResNet50  & 7.2  & 7.4 & 0.682 & 0.973 & - \\ 

Tang \etal \cite{handar}      & ResNet50 & 7.1  & 7.1 & 0.706 & 0.977 & 39.1 \\   
MobRecon\cite{mobrecon}     & DenseStack & 6.9  & 7.2 & 0.694 & 0.979 & 77 \\
CMR \cite{cmr_hand}                & ResNet50$^*$ & 6.9  & 7.0 & 0.715 & 0.977 & - \\ 
\textbf{Ours} & \textbf{FastViT-MA36} & \textbf{6.6} & \textbf{6.7} & \textbf{0.722} & \textbf{0.981} & \textbf{218} \\
\bottomrule

\end{tabular}
}
% \vspace{-1.5mm}
\caption{Results on the FreiHAND test dataset. FPS is computed on NVIDIA RTX-2080Ti under similar setting as MobRecon~\cite{mobrecon}. 
Performance of competing methods obtained from training exclusively on FreiHand dataset.
}
\label{tab:freihand}
% \vspace{-0.3cm}
\end{table}

\subsection{Semantic Segmentation and Object Detection}
For semantic segmentation, we validate the performance of our models on ADE20k~\cite{ade20k}. The dataset contains 20K training images and 2K validation images with 150 semantic categories. We train semantic segmentation models with Semantic FPN~\cite{Semantic_FPN} decoder. Models with Semantic FPN head use the same settings as~\cite{metaformer}. All models are initialized with pretrained weights from their corresponding image classification models. FLOPs and backbone latency are estimated on image crops of 512$\times$512. Due to higher resolution of input image, GPU latencies are estimated over a batch size of 2 in both Table~\ref{tab:segmentation} and Table~\ref{tab:detection}. In Table~\ref{tab:segmentation}, we compare our models with recent works. FastViT-MA36 model obtains 5.2\% higher mIoU than PoolFormer-M36 which has higher FLOPs, parameter count and latency on both desktop GPU and mobile device.

We train object detection on the MS-COCO~\cite{COCO} dataset with 80 classes containing 118K training and 5K validation images. In Table~\ref{tab:detection}, we compare our models with recent works. All the models are trained with 1x schedule following~\cite{metaformer} using a Mask-RCNN~\cite{Mask_RCNN} head. All models are initialized with pretrained weights from their corresponding image classification models. We show that our models achieve state-of-the-art performance under multiple latency regimes. FastViT-MA36 model has similar performance as CMT-S, while being 2.4$\times$ and 4.3$\times$ faster on desktop GPU and mobile device respectively.

\begin{table}[]
    \centering
    \resizebox{\linewidth}{!}{
    \begin{tabular}{l@{\hspace{0.5\tabcolsep}}|c@{\hspace{0.5\tabcolsep}}c|@{\hspace{0.5\tabcolsep}}c@{\hspace{0.5\tabcolsep}}c|c}
    \toprule
    \multirow{2.5}{*}{Backbone}    & GPU & Mobile    & \multicolumn{3}{c}{Semantic FPN 80k}  \\
    \cmidrule{4-6}
    & Latency  & Latency    & Param(M)   & FLOPs(G)   & mIoU(\%)  \\
    \midrule
    ResNet-50    & 2.7 &  8.7  & 28.5      &    46     & 36.7\\
    PoolFormer-S12\cite{metaformer}      &  9.7  &   6.2     & 15.7   & 31  & 37.2            \\
    \textbf{FastViT-SA12} & \textbf{2.5} & \textbf{5.6} &  14.1 & 29 & \textbf{38.0} \\
    \midrule
    ResNet-101 & 4.6 & 12.4 & 47.5 & 65 & 38.8 \\
    PVT-Small\cite{PVT_v1}   & - & - & 28.2 & 41 & 39.8 \\
    PoolFormer-S24\cite{metaformer}     & 18.9 &   10.7     & 23.2   & 48     & 40.3 \\
    \textbf{FastViT-SA24} & \textbf{4.4} & \textbf{9.3} & 23.8 & 37 & \textbf{41.0} \\
    \midrule
    PoolFormer-S36\cite{metaformer}     & 28.0 &   17.0     & 34.6   & 48     & 42.0   \\
    \textbf{FastViT-SA36}          & \textbf{6.1} &  \textbf{12.9} & 33.6      &   44      & \textbf{42.9}      \\ 
    \midrule
    
    PVT-Medium\cite{PVT_v1}   & - & - & 48.0 & 55 & 41.6 \\
    PoolFormer-M36\cite{metaformer} & 41.4 & 24.8 & 59.8 & 68 & 42.4 \\ 
    \textbf{FastViT-MA36}          & \textbf{8.2} & \textbf{16.3} & 45.7 & 53 & \textbf{44.6}      \\ 
    \bottomrule
    \end{tabular}}
    % \vspace{-1.5mm}
    \caption{Performance of different backbones on ADE20K semantic segmentation task. Following common convention, FLOPs and backbone latencies are measured on crops of 512$\times$512. 
    }\label{tab:segmentation}
    % \vspace{-2mm}
\end{table}

\begin{table}[t]
\scalebox{0.79}{
% 	\setlength\tabcolsep{4.0pt}
% 	\vspace{-0.3cm}
	\label{tab:detection}
	\centering
	\small 
	\begin{tabular}{l@{\hspace{0.3\tabcolsep}}|c@{\hspace{0.6\tabcolsep}}c@{\hspace{0.6\tabcolsep}}|c@{\hspace{0.6\tabcolsep}}c@{\hspace{0.6\tabcolsep}}c@{\hspace{0.6\tabcolsep}}c@{\hspace{0.6\tabcolsep}}c@{\hspace{0.6\tabcolsep}}c}
		\toprule
		\multirow{2}{*}{Backbone} & GPU & Mobile & \multirow{2}{*}{AP$^{\rm b}$} & \multirow{2}{*}{AP$_{50}^{\rm b}$} & \multirow{2}{*}{AP$_{75}^{\rm b}$} & \multirow{2}{*}{AP$^{\rm m}$} & \multirow{2}{*}{AP$_{50}^{\rm m}$} & \multirow{2}{*}{AP$_{75}^{\rm m}$} \\
		& Latency & Latency & & & & \\
		\midrule
		Poolfomer-S12\cite{metaformer} & 9.7 & 6.2 & 37.3 & 59.0 & 40.1 & 34.6 & 55.8 & 36.9 \\
		ResNet-50~\cite{resnet} & 2.7 & 8.7 & 38.0 & 58.6 & 41.4 & 34.4 & 55.1 & 36.7 \\
		\textbf{FastViT-SA12} & \textbf{2.5} & \textbf{5.6} & \textbf{38.9} & \textbf{60.5} & \textbf{42.2} &	\textbf{35.9} & \textbf{57.6} & \textbf{38.1} \\
		\midrule
		ResNet-101~\cite{resnet} & 4.6 & 12.4 & 40.0 & 60.5 & 44.0 & 36.1 & 57.5 & 38.6 \\
		Poolfomer-S24\cite{metaformer} & 18.9 & 10.7 & 40.1 & 62.2 & 43.4 & 37.0 & 59.1 & 39.6 \\
		PVT-S~\cite{PVT_v1} & - & - & 40.4 & 62.9 & 43.8 & 37.8 & 60.1 & 40.3 \\
		\textbf{FastViT-SA24} & \textbf{4.4} & \textbf{9.3} & \textbf{42.0} & \textbf{63.5} & \textbf{45.8} & \textbf{38.0} & \textbf{60.5} & \textbf{40.5} \\
		\midrule
		Swin-T~\cite{Swin} & - & - & 42.2 & 64.6 & 46.2 & 39.1 & 61.6 & 42.0 \\
		Twins-SVT-S~\cite{Twins} & - & - & 42.7 & 65.6 & 46.7 & 39.6 & 62.5 & 42.6 \\
		Twins-PCPVT-S~\cite{Twins} & - & 52.1 & 42.9 & 65.8 & 47.1 & 40.0 & 62.7 & 42.9 \\
		\textbf{FastViT-SA36} & \textbf{6.1} & \textbf{12.9} & \textbf{43.8} & \textbf{65.1} & \textbf{47.9} & \textbf{39.4} & \textbf{62.0} & \textbf{42.3} \\
		\midrule
		
		CMT-S~\cite{Cmt}  & 19.9 & 70.9 & 44.6 & \bf 66.8 & 48.9 & \bf 40.7 & \bf 63.9 & \bf 43.4 \\
        Poolfomer-S36\cite{metaformer} & 28.0 & 17.0 & 41.0 & 63.1 & 44.8 & 37.7 & 60.1 & 40.0 \\
        % \hline 
        \textbf{FastViT-MA36} & \textbf{8.2} & \textbf{16.3} & \textbf{45.1} & \textbf{66.8} & \textbf{49.5} & 40.5 & 63.8 & \textbf{43.4}  \\
		\bottomrule
	\end{tabular}
	}
% 	\vspace{-1.5mm}
    \caption{Results for object detection and instance segmentation on MS-COCO \texttt{val2017} split using Mask-RCNN~\cite{Mask_RCNN} framework using 1x training schedule, i.e. 12 epochs used for training the models. Backbone latencies are measured on crops of 512$\times$512. 
    }\label{tab:detection}
	% \vspace{-0.4cm}
\end{table}

%% file: conclusion.tex
\vspace{-2mm}
\section{Conclusion}
We have proposed a general purpose hybrid vision transformer that is highly efficient on multiple compute fabrics: mobile devices and desktop grade GPUs. 
Through structural reparameterization, our model incurs reduced memory access cost. This leads to significant improvements in runtime especially at higher resolutions. In addition, we propose further architectural changes that boosts performance on ImageNet classification task and other downstream tasks like object detection, semantic segmentation and 3D hand mesh estimation. We empirically show that our backbone is highly robust to out-of-distribution samples, while being significantly faster than competing robust models.

\label{sec:conclusion}

%% file: appendix.tex
\appendix
\section{Ablations}
\subsection{Architectural Choices}
The primary motivation behind the design choices made for FastViT is efficient mobile deployment. The cost of self-attention blocks is very high, especially when there are many tokens. In early stages (when the number of tokens are high), self attention can be replaced with efficient alternatives %(we use RepMixer in our design)
with small accuracy degradation but significantly lower latency. In Table 4 of main paper, we analyzed this for last two stages. In Table~\ref{tab:arch_choices_abl}, we present the full analysis for S12 architecture.

\begin{table}[]
\centering
\scalebox{0.7}{
\begin{tabular}{@{\hspace{0.5\tabcolsep}}l| @{\hspace{0.5\tabcolsep}}l| @{\hspace{0.5\tabcolsep}}c @{\hspace{1.0\tabcolsep}}c}
\toprule
\multirow{2}{*}{Ablation} & \multirow{2}{*}{Description} & Top-1 & Mobile       \\
                          & & Acc. & Latency (ms)  \\
\midrule
 - & Baseline & 79.8 & 1.4 \\
\midrule
 Normalization & BatchNorm $\rightarrow$ LayerNorm & 79.7 & 1.7 \\
\midrule
\multirow{2}{*}{Activation} & GELU $\rightarrow$ ReLU & 79.4 & 1.3 \\
                            & GELU $\rightarrow$ SiLU & 79.7 & 1.4 \\
\midrule
\multirow{4}{*}{Hybrid Stages} & [RepMix, RepMix, RepMix, SelfAttn.] & 80.6 & 1.6 \\
                               & [RepMix, RepMix, SelfAttn., SelfAttn.] & 81.2 & 3.7 \\
                              & [RepMix, SelfAttn., SelfAttn., SelfAttn.] & 81.5 & 5.0 \\
                              & [SelfAttn., SelfAttn., SelfAttn., SelfAttn.] & 82.0 & 11.7 \\
                            
\bottomrule
\end{tabular}
}
\caption{Ablation for FastViT-S12 on ImageNet-1K. All models are trained and benchmarked using the same settings described in main paper.}
\label{tab:arch_choices_abl}
\end{table}

\subsection{Large Kernel Convolutions}
In Table~\ref{tab:large_kernel_abl}, we ablate over various kernel sizes in FFN and patch embedding layers and report their Top-1 accuracy on ImageNet-1k along with mobile latency on iPhone 12 Pro. We observe that performance of the model stagnates beyond the kernel size of 7$\times$7, while the overall FLOPs, latency and parameter count increases. Hence, we use 7$\times$7 kernel size in our network architecture. 

\begin{table}[!htb]
\centering
\scalebox{0.90}{
\begin{tabular}{c|c|c|c|c}
\toprule
\multirow{2}{*}{Kernel Size}   & Params & FLOPs & Mobile & Top-1 \\
& (M) & (G) & Latency(ms) & (\%) \\
\midrule

3$\times$3   & 8.7 & 1.7 & 1.3 & 78.9 \\
5$\times$5   & 8.7 & 1.8 & 1.4 & 79.5 \\
7$\times$7   & 8.8 & 1.8 & 1.4 & 79.8 \\
9$\times$9   & 8.8 & 1.8 & 1.5 & 79.6 \\
11$\times$11 & 8.8 & 1.9 & 1.5 & 79.8 \\

\bottomrule
\end{tabular}
}
\caption{Top-1 accuracy on ImageNet-1k dataset for FastViT-S12 model with varied kernel sizes in FFN and patch embedding layers.}
\label{tab:large_kernel_abl}
\end{table}

\subsection{Training Time}
We provide a coarse ablation of this in Table 3 of main paper. In our model, we do not overparameterize every element of the architecture, especially the parameter dense blocks like ConvFFN. In fact, we do not obtain any improvement by overparameterizing ConvFFN layers, verified empirically in Table~\ref{tab:ttop_abl}. Since our model is partially overparameterized (only convolutional stem and patch embedding layers) during training, we do not see a significant degradation in train time as opposed to other train-time overparameterized models in literature which overparameterize all layers in a network. Note, all models were trained using the same hardware as described in Section 4.1 of main paper. 

\begin{table}[]
\centering
\scalebox{0.70}{
\begin{tabular}{@{\hspace{0.5\tabcolsep}}l| @{\hspace{0.5\tabcolsep}}l| @{\hspace{0.5\tabcolsep}}c @{\hspace{1.0\tabcolsep}}c}
\toprule
\multirow{2}{*}{Ablation} & \multirow{2}{*}{Description} & Top-1 & Train       \\
                                                      &  & Acc. & Time (hrs)  \\
\midrule
 No OverParam. & -                                                   & 80.0 & 31.3 \\
\midrule
\multirow{2}{*}{Train-Time OverParam.} & Stem + Patch Emb.           & 80.6 & 33.4 \\
                                       & Stem + Patch Emb. + ConvFFN & 80.6 & 40.1  \\

\bottomrule
\end{tabular}
}
\caption{Train-time overparameterization ablation for FastViT-SA12 on ImageNet-1K. Extension to Table 3 in main paper. Train time is wall clock time elapsed at the end of a training run.}
\label{tab:ttop_abl}

\end{table}

\section{Experiments}
\subsection{Benchmarking}
We follow the same protocol prescribed in~\cite{mobileone} while benchmarking the model on an iPhone 12 Pro mobile device. To benchmark the models on desktop-grade GPU NVIDIA RTX-2080Ti, we first warmup the device by running the forward pass of TensorRT model for 100 iterations and then benchmark the model over a 100 iterations. We report the median latency value from 100 estimates. For image classification models we use a batchsize of 8 (similar approach was adopted in~\cite{ShuffleNet_v2}) and a batchsize of 2 (due to GPU memory limits) for semantic segmentation and object detection models. Both mobile and GPU latency estimates can have a standard deviation of $\pm$0.1ms.

While benchmarking ConvNeXt~\cite{ConvNext} models on mobile device, we noticed the inefficiencies introduced by reshape ops causing increase in latency. While majority of hybrid models that use self-attention based token mixers require explicit reshaping of tensors. We can avoid this operation in ConvNeXt basic block by simply using a channel first implementation of LayerNorm and replacing \texttt{nn.Linear} layers with 1$\times$1 convolutions. This simple change results in significant improvement in runtime as shown in Table~\ref{tab:convnext_mod}.

\begin{table}[!htb]
\centering
\scalebox{0.8}{
\begin{tabular}{l|c|c}
\toprule
\multirow{2}{*}{Model}   & \multicolumn{2}{c}{Mobile Latency (ms)} \\
\cmidrule{2-3}
 & Before & After \\
\midrule
ConvNeXt-T & 33.5 & 3.7 \\
ConvNeXt-S & 66.4 & 5.4 \\
ConvNeXt-B & 89.1 & 8.4 \\
\bottomrule
\end{tabular}
}
\caption{Benchmarking ConvNeXt before and after modifications.}
\label{tab:convnext_mod}
\end{table}

\begin{table}[!htb]
\centering
\scalebox{0.68}{
\begin{tabular}{l|c|c}
\toprule
\multirow{2}{*}{Hyperparameter}   & Training & Fine-tuning \\
 & T8, T12, S12, SA12, SA24, SA36, MA36 & SA36, MA36 \\
\midrule
Stochastic depth rate & [0.0, 0.0, 0.0, 0.1, 0.1, 0.2, 0.35] & [0.2, 0.4] \\
Input resolution & 256$\times$256 & 384$\times$384 \\
Data augmentation & RandAugment & RandAugment \\
Mixup $\alpha$ & 0.8 & 0.8\\
CutMix $\alpha$ & 1.0 & 1.0 \\
Random erase prob. & 0.25 & 0.25 \\
Label smoothing & 0.1 & 0.1 \\
Train epochs & 300 & 30 \\
Warmup epochs & 5 & None \\
Batch size & 1024 & 1024 \\
Optimizer & AdamW & AdamW \\
Peak learning rate & 1e-3 & 5e-6 \\
LR. decay schedule & cosine & None \\
Weight decay rate & 0.05 & 1e-8 \\
Gradient clipping & None & None \\
EMA decay rate & 0.9995 & 0.9995 \\

\bottomrule

\end{tabular}
}
\caption{Training hyperparameters for ImageNet-1k experiments.}
\label{tab:hyperparams_inet1k}
\end{table}

\begin{figure*}[t]
    \centering
    \includegraphics[width=0.8\linewidth]{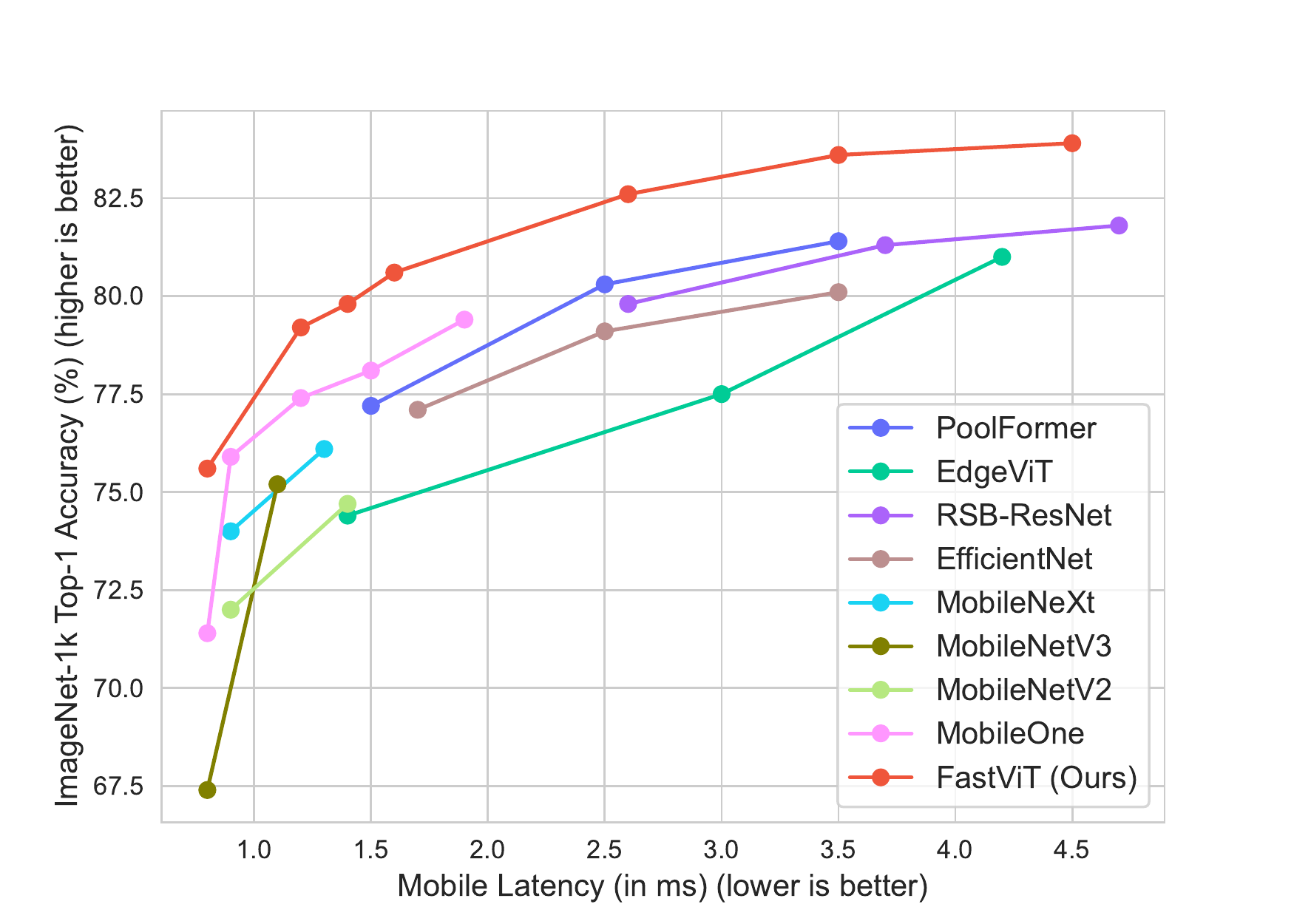}
    \caption{Accuracy vs. Mobile latency scaling curves of recent state-of-the-art Mobile Architectures and FastViT variants. The models are benchmarked on iPhone 12 Pro using the appropriate image sizes described in Table~\ref{tab:ImageNet_1K_mobile}. }
    \label{fig:mob_arch_plot}
\end{figure*}

\begin{table}[]
    \centering

    \scalebox{0.82}{
    \begin{tabular}{l| @{\hspace{0.5\tabcolsep}} c @{\hspace{1.0\tabcolsep}} c @{\hspace{1.0\tabcolsep}}c @{\hspace{1.0\tabcolsep}}c| @{\hspace{0.5\tabcolsep}}c}
    \toprule
    \multirow{2}{*}{Model}        & Eval    & Param     & FLOPs    & Mobile       & Top-1  \\ 
    % \cmidrule{5-6}
                                    & Image &      &       & Latency  &  Acc.      \\   
                    & Size & (M) & (G) & (ms) & (\%) \\
                                    
                                    \midrule

    MobileNetV3-S*\cite{Mobilenet_v3} & 224 & 2.5 & 0.06 & \textbf{0.8} & 67.4 \\
    MobileOne-S0\cite{mobileone} & 224 & 2.1 & 0.3 & \textbf{0.8} & 71.4 \\ 
    MobileNetV2-x1.0\cite{MobileNet_v2} & 224 & 3.4 & 0.3 & 0.9 & 72.0 \\
    DeiT-Ti\cite{deit}  & 224 & 5.7 & 1.3 & 4.8 & 72.2 \\
    MobileNeXt-x1.0\cite{mobilenext_eccv2020} & 224 & 3.4 & 0.3 & 0.9 & 74.0 \\
    EdgeViT-XXS\cite{edgevit} & 224 & 4.1 & 0.6 & 1.4 & 74.4 \\
    MNASNet-A1\cite{mnasnet_cvpr} & 224 & 3.9 & 0.3 & 1.0 & 75.2 \\
    \textbf{FastViT-T8} & 256 & 3.6 & 0.7 & \textbf{0.8} & \textbf{75.6} \\ 

    \midrule

    MobileNetV2-x1.4\cite{MobileNet_v2} & 224 & 6.9 & 0.6 & 1.4 & 74.7 \\
    MobileNetV3-L\cite{Mobilenet_v3} & 224 & 5.4 & 0.2 & 1.1 & 75.2 \\
    MobileNeXt-x1.4\cite{mobilenext_eccv2020}  & 224 & 6.1 & 0.6 & 1.3 & 76.1 \\
    EfficientNet-B0\cite{EfficientNet} & 224 & 5.3 & 0.4 & 1.7 & 77.1 \\
    EdgeViT-XS\cite{edgevit} & 224 & 6.7 & 1.1 & 3.0 & 77.5 \\
    MobileOne-S3\cite{mobileone} & 224 & 10.1 & 1.8  & 1.5 & 78.1 \\ 
    CMT-$\text{T}^*$\cite{Cmt}  & 160 & 9.5 & 0.6  & 3.8 & \textbf{79.1} \\
    EfficientNet-B1\cite{EfficientNet} & 256 & 7.8 & 0.7 & 2.5 & \textbf{79.1} \\
    \textbf{FastViT-T12} & 256 & 6.8 & 1.4 &  \textbf{1.2} & \textbf{79.1} \\

    \midrule
    MobileOne-S4\cite{mobileone} & 224 & 14.8 & 2.9  & 1.9 & 79.4 \\
    DeiT-S\cite{deit}  & 224 & 22.0 & 4.6 & 5.3 & \textbf{79.8} \\
    \textbf{FastViT-S12} & 256 & 8.8 & 1.8 &  \textbf{1.4} & \textbf{79.8} \\
 
    \bottomrule
    \end{tabular}
    
    }
        \caption{Comparison of different state-of-the-art Mobile architectures on ImageNet-1k classification. HardSwish is not well supported by Core ML, $*$ denotes we replace it with GELU for fair comparison.}
    \label{tab:ImageNet_1K_mobile}
\end{table}

\subsection{Image Classification}
We provide hyperparameters used for training models on ImageNet-1k dataset reported in Table 5 in main paper. Models are trained at resolution 256$\times$256 and fine-tuned for 384$\times$384. 
We follow the same training setup as~\cite{metaformer, deit}. The hyperparameters used for all FastViT variants are listed in Table~\ref{tab:hyperparams_inet1k}. For distillation experiments, we use RegNetY-16GF~\cite{RegNet} as the teacher model similar to~\cite{deit}. Additional hyperparameters are same as our image classification training procedure and are listed in Table~\ref{tab:hyperparams_inet1k}. When trained using different seeds, results are within $\pm$0.2\% in Top-1 accuracy.

\subsection{Comparison with Mobile Architectures}
We compare our model against highly efficient mobile architectures in Table~\ref{tab:ImageNet_1K_mobile} and in Figure~\ref{fig:mob_arch_plot}. Our model outperforms recent state-of-the-art MobileOne~\cite{mobileone} architecture which is purely convolutional. Our model also outperforms EdgeViT~\cite{edgevit}, which is a recent state-of-the-art light-weight ViT architecture.

\begin{figure*}[t]
    \centering
    \includegraphics[width=0.85\linewidth]{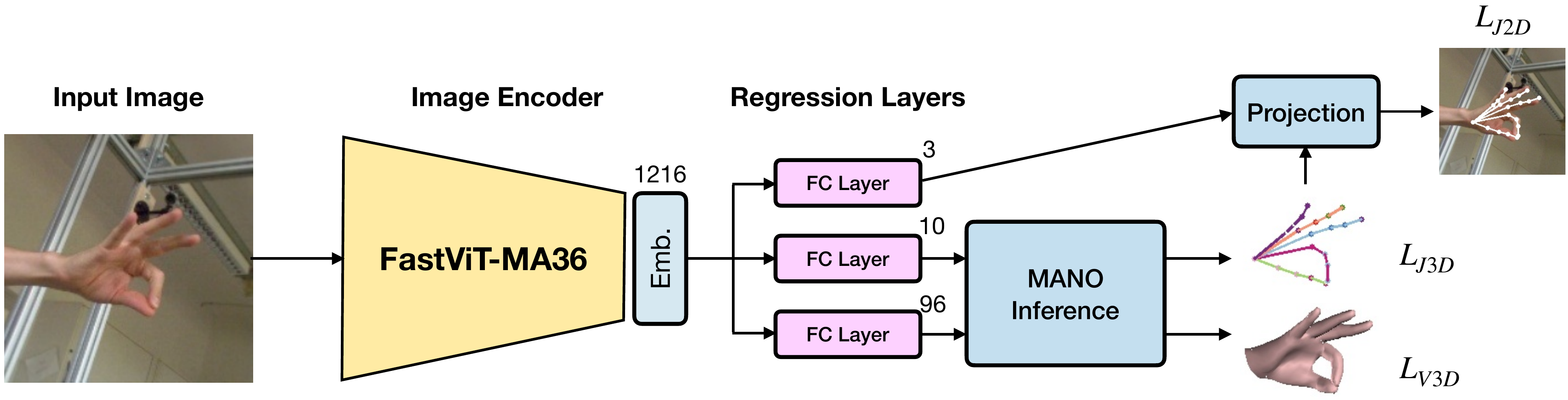}
    \caption{Overview of 3d hand mesh estimation framework.}
    \label{fig:hand_mesh_arch}
\end{figure*}

\begin{figure*}[t]
    \centering
    \includegraphics[width=0.95\linewidth]{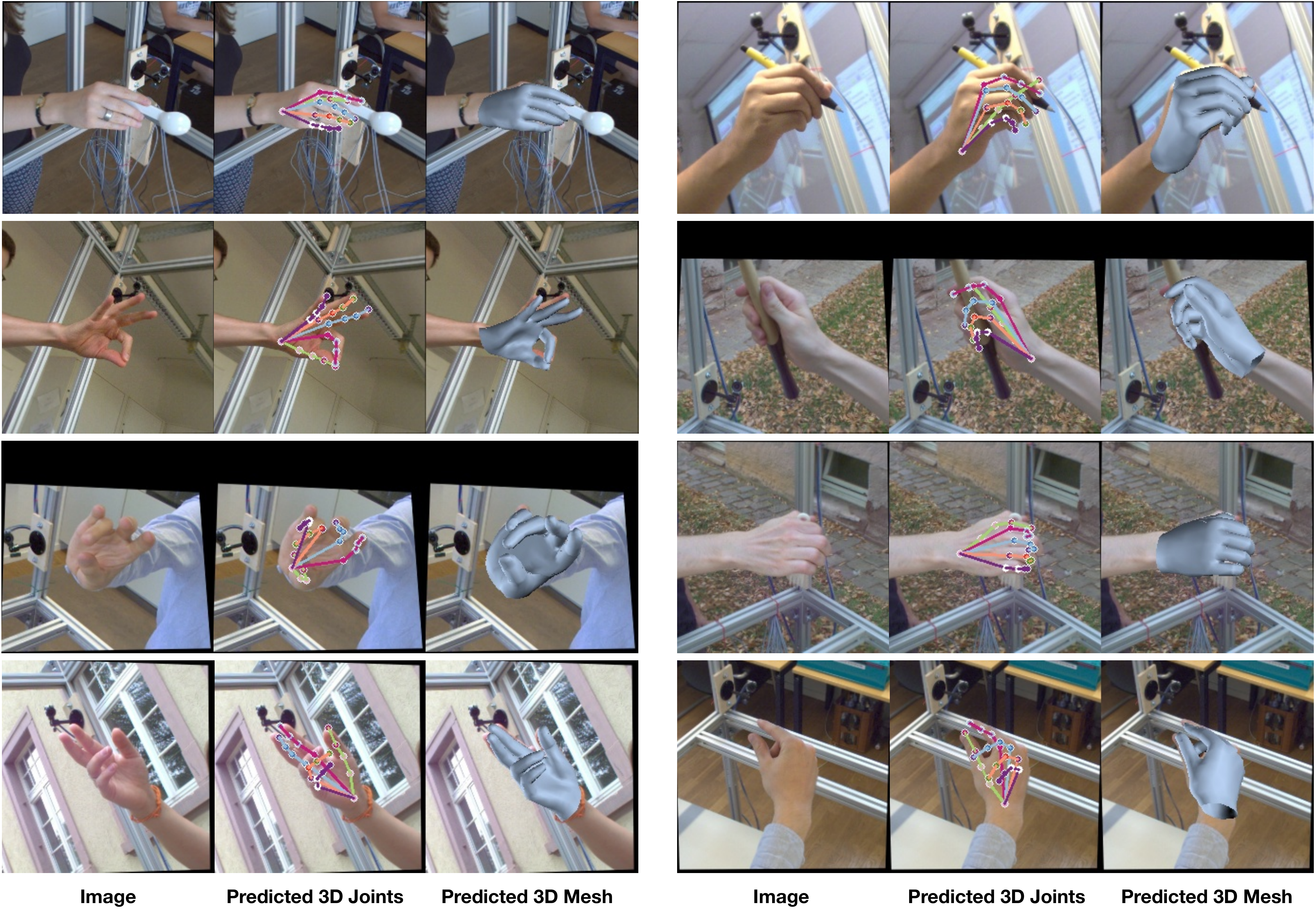}
    \caption{Qualitative results from our framework on FreiHand test set. 3D predictions are projected on to the image using weak perspective camera model, parameters for this camera model is also predicted by the model.}
    \label{fig:hand_mesh_results}
\end{figure*}

\subsection{Model Scaling} 
In this work, we sample architectures that are smaller than 50M parameters for efficient deployment. Similar to the Swin-T~\cite{Swin} variant, we use a stage compute ratio of 1:1:3:1 most of our variants and for the smallest variant we use 1:1:2:1. For our models, width doubles at each new stage. We use configurations of [48, 96, 192, 384], [64, 128, 256, 512] and [76, 152, 304, 608] in this paper. The tiny(``T") variants use MLP expansion ratio of 3. Rest of the variants use an MLP expansion ratio of 4.

\subsection{3D Hand mesh estimation}

\paragraph{Architecture}
As shown in Figure~\ref{fig:hand_mesh_arch}, our model uses simple regression layers to regress weak perspective camera, pose and shape parameters of the MANO model~\cite{manomodel}. These layers are single fully connected layers unlike deep regression layers used in~\cite{mobilehand}. We regress 6D rotations~\cite{6drot} for all joints in MANO model. There are 3 losses to minimize in our framework, $L_{V3D}$, 3D vertex loss between predicted mesh and ground truth mesh. $L_{J3D}$, 3D joint loss between predicted 3D joints and ground truth 3D joints. $L_{J2D}$, 2D joint loss between projected 3D joints and ground 2D keypoints.

\vspace{-3mm}
\paragraph{Setup}
We train our model on FreiHand~\cite{freihand} dataset, that contains 130,240 training images and 3,960 test images. Following METRO~\cite{metro_hand}, we train our model on 224$\times$224 images from the dataset using Adam optimizer. The models are trained for 200 epochs, with an initial learning rate of 1e-4 and decayed by a factor of 10 after 100 epochs. We initialize the backbone with weights obtained by pretraining on ImageNet-1k dataset. The weighting for all the losses in our framework is set to 1.0.

\paragraph{Results}
Figure~\ref{fig:hand_mesh_results} shows qualitative results from our framework on FreiHand test set. Even though our model is simple, it can model complicated gestures. Our model predicts reliable poses even in the presence of occlusion from hand-held objects.